\documentclass[headings=standardclasses]{scrartcl}

\usepackage{geometry}
\usepackage[utf8]{inputenc}
\usepackage[T1]{fontenc}
\usepackage[onehalfspacing]{setspace}
\usepackage{charter}
\usepackage[charter]{mathdesign}
\usepackage{microtype}
\usepackage{amsmath}
\usepackage{amsthm}
\usepackage{tabulary}
\usepackage{longtable}
\usepackage{booktabs} 
\usepackage{ragged2e}

\usepackage{siunitx}
\usepackage{bm}
\usepackage[pdftex]{graphicx}
\usepackage{rotating}
\usepackage{pdfpages}
\usepackage{pdflscape}
\usepackage{color}
\usepackage{array}
\usepackage[font=bf]{caption}
\usepackage{subcaption}
\usepackage{pgfplots}
\usepackage{pgfplotstable}
\usepackage{float}
\usepackage[title]{appendix}
\pgfplotsset{compat=1.9}
\usepackage{url}
\usepackage[section]{placeins}
\usepackage{natbib}

\usepackage[boxed]{algorithm2e}
\SetKwFunction{KwInit}{Initialisation}
\SetKwComment{Comment}{\% }{}

\makeatletter
\renewcommand*\env@matrix[1][c]{\hskip -\arraycolsep
  \let\@ifnextchar\new@ifnextchar
  \array{*\c@MaxMatrixCols #1}}
\makeatother

\usepackage{xcolor}
\usepackage{soul}

\usepackage{verbatim}

\begin{document}
\begin{spacing}{1.2}
\begin{flushleft}
\Large \textbf{P{\'o}lygamma Data Augmentation to address Non-conjugacy in the Bayesian Estimation of Mixed Multinomial Logit Models} \\
\vspace{\baselineskip}
\normalsize
April 13, 2019 \\
\vspace{\baselineskip}
Prateek Bansal\textsuperscript{*} \\
School of Civil and Environmental Engineering \\
Cornell University, United States \\
pb422@cornell.edu \\
\vspace{\baselineskip}
Rico Krueger\textsuperscript{*} \\
Research Centre for Integrated Transport Innovation, School of Civil and Environmental Engineering, UNSW Australia, Sydney NSW 2052, Australia \\
r.krueger@student.unsw.edu.au \\
\vspace{\baselineskip}
Michel Bierlaire \\
Transport and Mobility Laboratory, School of Architecture, Civil and Environmental Engineering, Ecole Polytechnique Fédérale de Lausanne, Station 18, Lausanne 1015, Switzerland \\
michel.bierlaire@epfl.ch \\
\vspace{\baselineskip}
Ricardo A. Daziano \\
School of Civil and Environmental Engineering \\
Cornell University, United States  \\
daziano@cornell.edu \\
\vspace{\baselineskip}
Taha H. Rashidi  \\
Research Centre for Integrated Transport Innovation, School of Civil and Environmental Engineering, UNSW Australia, Sydney NSW 2052, Australia\\
rashidi@unsw.edu.au\\
\vspace{\baselineskip}
\textsuperscript{*} These authors contributed equally to this work.
\end{flushleft}
\end{spacing}

\newpage
\section*{Abstract}
The standard Gibbs sampler of Mixed Multinomial Logit (MMNL) models involves sampling from conditional densities of utility parameters using Metropolis-Hastings (MH) algorithm due to unavailability of conjugate prior for logit kernel. To address this non-conjugacy concern, we propose the application of P{\'o}lygamma data augmentation (PG-DA) technique for the MMNL estimation. The posterior estimates of the augmented and the default Gibbs sampler are similar for two-alternative scenario (binary choice), but we encounter empirical identification issues in the case of more alternatives ($J \geq 3$).

\section{Mixed multinomial logit model} \label{sec:MMNL}
The mixed multinomial logit (MMNL) model \citep{mcfadden2000mixed} is established as follows: We consider a standard discrete choice setup, in which on choice occasion $t \in \{1, \ldots T \}$, a decision-maker $n \in \{1, \ldots N \}$ derives utility $U_{ntj} = V(\boldsymbol{X}_{ntj}, \boldsymbol{\Gamma}_{n}) + \epsilon_{ntj}$ from alternative $j \in \{1, \ldots J \}$. Here, $V()$ denotes the representative utility, $\boldsymbol{X}_{ntj}$ is a row-vector of covariates, $\boldsymbol{\Gamma}_{n}$ is a collection of taste parameters, and $\epsilon_{ntj}$ is a stochastic disturbance. The assumption $\epsilon_{ntj} \sim \text{Gumbel}(0,1)$ leads to a multinomial logit (MNL) kernel such that the probability that decision-maker $n$ chooses alternative $j$ on choice occasion $t$ is 
\begin{equation}
P(y_{nt} = j \vert \boldsymbol{X}_{ntj}, \boldsymbol{\Gamma}_{n}) = \frac{\exp \left \{ V (\boldsymbol{X}_{ntj}, \boldsymbol{\Gamma}_{n}) \right \}}{\sum_{k = 1}^J \exp \left \{ V (\boldsymbol{X}_{ntk}, \boldsymbol{\Gamma}_{n}) \right \}},
\end{equation} 
where $y_{nt}$ captures the observed choice. The choice probability can be iterated over choice scenarios to obtain the probability of observing a decision-maker's sequence of choices $\boldsymbol{y}_{n}$:
\begin{equation}
P(\boldsymbol{y}_{n} \vert \boldsymbol{X}_{n},  \boldsymbol{\Gamma}_{n}) = \prod_{t = 1}^{T} P(y_{nt} = j \vert \boldsymbol{X}_{nt},  \boldsymbol{\Gamma}_{n}).
\end{equation}

We consider a general utility specification under which tastes $\boldsymbol{\Gamma}_{n}$ are partitioned into fixed taste parameters $\boldsymbol{\alpha}$, which are invariant across decision-makers, and random taste parameters $\boldsymbol{\beta}_{n}$, which are individual-specific, such that $\boldsymbol{\Gamma}_{n} = \begin{bmatrix} \boldsymbol{\alpha}^{\top} & \boldsymbol{\beta}_{n}^{\top} \end{bmatrix} ^{\top}$, whereby $\boldsymbol{\alpha}$ and $\boldsymbol{\beta}_{n}$ are vectors of lengths $L$ and $K$, respectively. Analogously, the row-vector of covariates $\boldsymbol{X}_{ntj}$ is partitioned into attributes $\boldsymbol{X}_{ntj,F}$, which pertain to the fixed parameters $\boldsymbol{\alpha}$, as well as into attributes $\boldsymbol{X}_{ntj,R}$, which pertain to the individual-specific parameters $\boldsymbol{\beta}_{n}$, such that $\boldsymbol{X}_{ntj} = \begin{bmatrix} \boldsymbol{X}_{ntj,F} & \boldsymbol{X}_{ntj,R} \end{bmatrix}$. For simplicity, we assume that the representative utility is linear-in-parameters, i.e. 
\begin{equation}
V (\boldsymbol{X}_{ntj}, \boldsymbol{\Gamma}_{n}) = \boldsymbol{X}_{ntj} \boldsymbol{\Gamma}_{n} = \boldsymbol{X}_{ntj,F} \boldsymbol{\alpha} + \boldsymbol{X}_{ntj,R} \boldsymbol{\beta}_{n}.
\end{equation}

The distribution of tastes $\boldsymbol{\beta}_{1:N}$ is assumed to be multivariate normal, i.e. $\boldsymbol{\beta}_{n} \sim \text{N}(\boldsymbol{\zeta}, \boldsymbol{\Omega})$ for $n = 1, \dots, N$, where $\boldsymbol{\zeta}$ is a mean vector and $\boldsymbol{\Omega}$ is a covariance matrix. In a fully Bayesian setup, the invariant (across individuals) parameters $\boldsymbol{\alpha}$, $\boldsymbol{\zeta}$, $\boldsymbol{\Omega}$ are also considered to be random parameters and are thus given priors. We use normal priors for the fixed parameters $\boldsymbol{\alpha}$ and for the mean vector $\boldsymbol{\zeta}$. Following \citet{tan2017stochastic} and \citet{akinc2018bayesian}, we employ Huang's half-t prior \citep{huang2013Simple} for covariance matrix $\boldsymbol{\Omega}$, as this prior specification exhibits superior noninformativity properties compared to other prior specifications for covariance matrices. In particular, \citep{akinc2018bayesian} show that Huang's half-t prior outperforms the inverse Wishart prior, which is often employed in fully Bayesian specifications of MMNL models \citep[e.g.][]{train2009discrete}, in terms of parameter recovery.

Stated succinctly, the generative process of the fully Bayesian MMNL model is:
\begin{align}
& \boldsymbol{\alpha} \lvert \boldsymbol{\lambda}_{0},\boldsymbol{\Xi}_{0} \sim \text{N}(\boldsymbol{\lambda}_{0},\boldsymbol{\Xi}_{0})\\
& \boldsymbol{\zeta} \vert \boldsymbol{\mu}_{0},\boldsymbol{\Sigma}_{0} \sim \text{N}(\boldsymbol{\mu}_{0},\boldsymbol{\Sigma}_{0}) \\
& a_{k} \lvert A_{k} \sim \text{Gamma}\left( \frac{1}{2}, \frac{1}{A_{k}^{2}} \right), & & k = 1,\dots,K,  \label{eq_gamma_a} \\
& \boldsymbol{\Omega} \vert \nu, \boldsymbol{a}\sim \text{IW}\left(\nu + K - 1, 2\nu \text{diag}(\boldsymbol{a}) \right),  \quad \boldsymbol{a} = \begin{bmatrix} a_{1} & \dots & a_{K} \end{bmatrix}^{\top} \label{eq_iv_Omega} \\
& \boldsymbol{\beta}_{n} \vert \boldsymbol{\zeta}, \boldsymbol{\Omega} \sim \text{N}(\boldsymbol{\zeta}, \boldsymbol{\Omega}), & & n = 1,\dots,N, \\
& y_{nt} \vert \boldsymbol{\alpha}, \boldsymbol{\beta}_{n}, \boldsymbol{X}_{nt} \sim \text{MNL}(\boldsymbol{\alpha}, \boldsymbol{\beta}_{n}, \boldsymbol{X}_{nt}), & & n = 1,\dots,N,  \ t = 1,\dots,T,
\end{align} 
where (\ref{eq_gamma_a}) and (\ref{eq_iv_Omega})  induce Huang's half-t prior \citep{huang2013Simple}. $\{ \boldsymbol{\lambda}_{0}, \boldsymbol{\Xi}_{0}, \boldsymbol{\mu}_{0}, \boldsymbol{\Sigma}_{0}, \nu, A_{1:K} \}$ are known hyper-parameters, and $\boldsymbol{\theta} = \{ \boldsymbol{\alpha}, \boldsymbol{\zeta}, \boldsymbol{\Omega},  \boldsymbol{a}, \boldsymbol{\beta}_{1:N}\}$ is a collection of model parameters whose posterior distribution we wish to estimate.

\section{P{\'o}lya--Gamma data augmentation} 
The default Gibbs sampler for posterior inference in MMNL models involves Metropolis steps to take draws from conditional densities of the utility parameters ($\bm{\beta}_{n}$ and $\bm{\alpha}$) due to the unavailability of a conjugate prior for the MNL kernel. MCMC estimation of binary and multinomial logistic regression models encounters a similar issue of non-conjugacy. P{\'o}lya-Gamma data augmentation (PG-DA) is the state-of-the-art technique to handle non-conjugacy in  MCMC estimation of binary logistic regression models \citep{polson2013bayesian}. PG-DA augments the Gibbs sampler by introducing an additional P{\'o}lya-Gamma distributed latent variable, which circumvents the need of the Metropolis algorithm by ensuring conjugate updates. \citet{polson2013bayesian} also extend PG-DA to the multinomial logistic regression model. Yet, this extension requires all utility (or link function) parameters to be alternative-specific. We use the same idea in deriving a PG-DA-based Gibbs sampler for MMNL, but we have to consider the same restriction on utility specification, i.e. replace $\bm{\Gamma}_{n}$ by $\bm{\Gamma}_{nj}$. 

\subsection{Augmented Gibbs Sampler}
The representative utility is: $V_{ntj} = \bm{X}_{ntj} \bm{\Gamma}_{nj} =\bm{X}_{ntj,F} \bm{\alpha}_{j} + \bm{X}_{ntj,R} \bm{\beta}_{nj}$, where $\bm{\beta}_{nj} \sim \text{N}(\bm{\zeta}_j,\bm{\Omega})$. The hyper-parameters remain the same, but the model parameters are  $\bm{\theta} = \{ \bm{\alpha}_{1:J}, \bm{\zeta}_{1:J}, \bm{\Omega},  a_{1:K}, \bm{\beta}_{\{1:N,1:J\}} \}$. 
Adhering to the original notation, we can write the joint distribution of the data and the model parameters:
\begin{equation}
\begin{split}
P & (\bm{y}_{1:N}, \bm{\theta}) =  \\
& P(\bm{\Omega} \vert \omega, \bm{B}) \prod_{n=1}^{N} P(\boldsymbol{y}_{n} \vert \boldsymbol{X}_{n}, \bm{\Gamma}_{n})
\prod_{j=1}^{J} P(\bm{\alpha}_{j} \vert \bm{\lambda}_{0},\bm{\Xi}_{0})
P(\bm{\zeta}_{j} \vert \bm{\mu}_{0},\bm{\Sigma}_{0})
\prod_{k=1}^{K} P(a_{k} \lvert  s,  r_{k})
\prod_{n=1}^{N}\prod_{j=1}^{J} P(\bm{\beta}_{nj} \vert \bm{\zeta}_{j}, \bm{\Omega}).
\end{split}
\end{equation}

Algorithm \ref{alg:polyagama} presents the augmented Gibbs sampler for the MMNL model. The conditional densities of $a_{1:K}$, $\bm{\Omega}$, and $\bm{\zeta}_{1:J}$ are similar to those of the Allenby-Train procedure \citep{akinc2018bayesian}. The next subsection details the derivation of conditional densities of $\bm{\beta}_{\{1:N,1:J\}}$ and $\bm{\alpha}_{1:J}$. 
	\begin{algorithm}[h]
		\SetAlgoLined
        \For{(iteration in 1 to max-iteration)}{
          Sample $a_{k}$ for $\forall k$ from $\text{Gamma}\left ( \frac{\nu + K}{2},  \frac{1}{A_{k}^{2}} + \nu \left ( \bm{\Omega}^{-1} \right )_{kk} \right )$  \; 
		  Sample $\bm{\Omega}$ from $\text{IW} \left (\nu + NJ + K - 1, 2 \nu \text{diag}(\bm{a})  + \sum_{n=1}^{N}\sum_{j=1}^{J} (\bm{\beta}_{nj} -  \bm{\zeta}_{j}) (\bm{\beta}_{nj} -  \bm{\zeta}_{j})^{\top} \right) $ \;
		\For{($i$ in $1$ to $J$)}{
		     Sample $\bm{\zeta}_{i}$ from $\text{N}\left ( \frac{1}{N} \sum_{n=1}^{N} \bm{\beta}_{ni}, \frac{\bm{\Omega}}{N} \right )$\;
            Sample $\bm{\beta}_{ni}$ for $\forall n$ using equation \ref{eq:pgbeta} \;   
            Update $\eta_{nti}$ and $L_{nti}$ for $\forall nt,i$ using equation \ref{eq:eta} \;
            Sample $\bm{\alpha}_i$ using equation \ref{eq:pgalpha}\; 
            Update $\eta_{nti}$ and $L_{nti}$ for $\forall nt,i$ using equation \ref{eq:eta}\;
            Sample $\phi_{nti}$ for $\forall n, t$ from $\text{PG}(1,\eta_{nti})$ \;
		}}
		\caption{P$\acute{o}$lya-Gamma augmented Gibbs sampler for the MMNL Model} \label{alg:polyagama}
	\end{algorithm}

\subsection{Conditional distributions of $\bm{\beta}_{nj}$ and $\bm{\alpha}_j$} 
Using \cite{holmes2006bayesian}, we can convert the multinomial logit likelihood expression to the binary logit likelihood: 
\begin{equation}
\begin{split}
 P(\bm{\beta}_{nj} \lvert \bm{y}_{1:N},\bm{\theta}_{-\bm{\beta}_{nj}}) & \propto P(\bm{\beta}_{nj} \lvert \bm{\zeta}_{j},\bm{\Omega}) \prod_{t=1}^T \left( \frac{\exp(\eta_{ntj})}{1+\exp(\eta_{ntj})}\right)^{y_{ntj}} \left( \frac{1}{1+\exp(\eta_{ntj})}\right)^{(1-y_{ntj})}\\
 & \propto P(\bm{\beta}_{nj} \lvert \bm{\zeta}_{j},\bm{\Omega}) \prod_{t=1}^T\left[\frac{\exp(\eta_{ntj})^{y_{ntj}}}{1+\exp(\eta_{ntj})}\right]
\end{split}
\end{equation}
where $\bm{\theta}_{-\bm{\beta}_{nj}}$ is a resulting parameter vector after removing $\bm{\beta}_{nj}$ and 
\begin{equation}\label{eq:eta}
\eta_{ntj} = V_{ntj} - L_{ntj};  \; L_{ntj} = \ln\left(\sum_{k=1, k\neq j}^J \exp(V_{ntk}) \right)
\end{equation}
We now introduce a P{\'o}lya--Gamma distributed auxiliary variable $\phi_{ntk} \sim \text{PG}(1,0) \; \forall n,t,k$ and $\kappa_{ntk} = y_{ntk}- \frac{1}{2}$. Now consider the identity derived by \cite{polson2013bayesian}:
\begin{equation}
    \frac{\exp(\eta_{ntk})^{y_{ntk}}}{1+\exp(\eta_{ntk})} = \frac{\exp(\kappa_{ntk}\eta_{ntk})}{2} \int_{0}^{\infty} \exp\left( -\frac{\eta_{ntk}^2 \phi_{ntk}}{2} \right) P(\phi_{ntk}) d\phi_{ntk}
\end{equation}
The conditional density of $\bm{\beta}_{nj}$ is:
\begin{equation}
    \begin{split}
    P(\bm{\beta}_{nj} \lvert \bm{y}_{1:N},\bm{\theta}_{-\bm{\beta}_{nj}}, \bm{\phi}) & \propto \exp\left(-\frac{1}{2}(\bm{\beta}_{nj} - \bm{\zeta}_{j})^{\top} \bm{\Omega}^{-1} (\bm{\beta}_{nj} - \bm{\zeta}_{j}) \right) \dots \\
    & \dots \prod_{t=1}^T \exp\left(\kappa_{ntj}\bm{X}_{ntj,R} \bm{\beta}_{nj} - \frac{\Big(\bm{X}_{ntj,F} \bm{\alpha}_{j}+ \bm{X}_{ntj,R} \bm{\beta}_{nj} - L_{ntj}\Big)^2\phi_{ntj}}{2} \right)
    \end{split}
\end{equation}
The conditional distribution of $\bm{\beta}_{nj}$ is:
\begin{equation}\label{eq:pgbeta}
    \begin{split}
    \bm{\beta}_{nj} & \lvert \bm{y}_{1:N},\bm{\theta}_{-\bm{\beta}_{nj}}, \bm{\phi} \\ 
    \sim & \text{N}\Bigg(\left[\bm{\Omega}^{-1} + \sum_{t=1}^T \phi_{ntj}\bm{X}_{ntj,R}^{\top}\bm{X}_{ntj,R}\right]^{-1} \\ &
    \left[\bm{\Omega}^{-1}\bm{\zeta}_{j} + \sum_{t=1}^T  \bm{X}_{ntj,R}^{\top} \left[\kappa_{ntj} - \phi_{ntj}(\bm{X}_{ntj,F} \bm{\alpha}_{j}- L_{ntj})\right] \right], \\
    & \left[\bm{\Omega}^{-1} + \sum_{t=1}^T \phi_{ntj}\bm{X}_{ntj,R}^{\top}\bm{X}_{ntj,R}\right]^{-1} \Bigg)
    \end{split}
\end{equation}
The conditional density of $\bm{\alpha}_{j}$ can be derived similarly:
\begin{equation}\label{eq:pgalpha}
    \begin{split}
    \bm{\alpha}_{j} & \lvert \bm{y}_{1:N},\bm{\theta}_{-\bm{\alpha}_{j}}, \bm{\phi}  \\
    \sim & \text{N}\Bigg(\left[\bm{\Xi}_{0}^{-1} + \sum_{n=1}^N\sum_{t=1}^T  \phi_{ntj} \bm{X}_{ntj,F}^{\top} \bm{X}_{ntj,F}\right]^{-1} \\&
    \left[\bm{\Xi}_{0}^{-1}\bm{\lambda}_{0} + \sum_{n=1}^N \sum_{t=1}^T  \bm{X}_{ntj,F}^{\top}\left[\kappa_{ntj} - \phi_{ntj}(\bm{X}_{ntj,R} \bm{\beta}_{nj}- L_{ntj})\right] \right], \\
    & \left[\bm{\Xi}_{0}^{-1} + \sum_{n=1}^N\sum_{t=1}^T \phi_{ntj}\bm{X}_{ntj,F}^{\top}\bm{X}_{ntj,F}\right]^{-1} \Bigg)
 \end{split}
\end{equation}

\subsection{Discussion}
We test the performance of the PG-DA-based Gibbs sampler against the Metropolis-based Gibbs sampler in a Monte Carlo study. We first consider the MNL model ($\bm{\Gamma}_{nj} = \bm{\alpha}_{j}$), where both samplers perform equally well. For MMNL model, the posterior estimates of the proposed PG-DA approach and the default Gibbs sampler are similar for $J=2$, but we encounter an explosion of the conditional distribution parameters in the case of more alternatives ($J \geq 3$). Results for $J=2$ and MATLAB code is available upon request. 

This appears to be an issue of empirical identification because of too many model parameters. Before the PG-DA sampler diverges, representative utilities are either very small or very large in magnitude for all alternatives across all observations. Therefore, instead of the actual magnitude of utilities, their comparative scales determine the choice probabilities. Thus, the algorithm might have a tendency to increase the relative scale of the latent utilities by increasing the scale of the parameters. 

In fact, prior to divergence the probability estimates of the chosen and non-chosen alternatives are close to one and zero, respectively. We speculate that such behavior might be a consequence of too many model parameters, which might allow the algorithm to find a parameter configuration that can fit the data very well (in terms of choice probabilities). Once the algorithm finds that configuration, it starts to increase the relative scale between the utilities (thus allowing the chosen alternatives to have probability close to one), causing the parameter explosion. 

As future research, stick-breaking constructions can be explored to adopt PG-DA in MCMC estimation of MMNL while keeping a parsimonious model specification, i.e. with generic utility parameters \citep{linderman2015dependent,zhang2017permuted}. However, before adopting these constructions, consistency with microeconomic theory needs to be established first.  

\bibliographystyle{apalike}
\bibliography{bibliography}

\end{document}